\documentclass[11pt,letterpaper]{article}
\usepackage{emnlp2017}
\usepackage{times}
\usepackage{latexsym}

\usepackage{url}
\usepackage[utf8]{inputenc}
\usepackage{amsmath}
\usepackage{multirow}
\usepackage{enumitem}
\usepackage{graphicx}
\usepackage{pgf}
\usepackage{tikz}
\usepackage{subfig}
\usepackage{array}
\usepackage{placeins}

\newcolumntype{L}{>{\arraybackslash}m{13cm}}

\emnlpfinalcopy

\def\emnlppaperid{754}

\title{Unsupervised Pretraining for Sequence to Sequence Learning}

\author{Prajit Ramachandran \and Peter J. Liu \and Quoc V. Le \\
Google Brain \\
  {\tt \{prajit, peterjliu, qvl\}@google.com}}

\date{}

\begin{document}

\maketitle

\begin{abstract}
This work presents a general unsupervised learning method to improve
the accuracy of sequence to sequence (seq2seq) models. In our method, the
weights of the encoder and decoder of a seq2seq model are initialized
with the pretrained weights of two language models and then 
fine-tuned with labeled data. We apply this method to
challenging benchmarks in machine translation and abstractive
summarization and find that it significantly improves the subsequent
supervised models.  Our main result is that pretraining improves the generalization of seq2seq models.
We achieve state-of-the-art results on the WMT
English$\rightarrow$German task, surpassing a range of methods using
both phrase-based machine translation and neural machine
translation. Our method achieves a significant improvement of 1.3 BLEU from the
previous best models on both WMT'14 and WMT'15
English$\rightarrow$German. We also conduct human evaluations on abstractive summarization and find that our method outperforms
a purely supervised learning baseline in a statistically significant manner.
\end{abstract}

\section{Introduction}

\begin{figure*}[h!]
\begin{center}
\scalebox{0.3}{\input{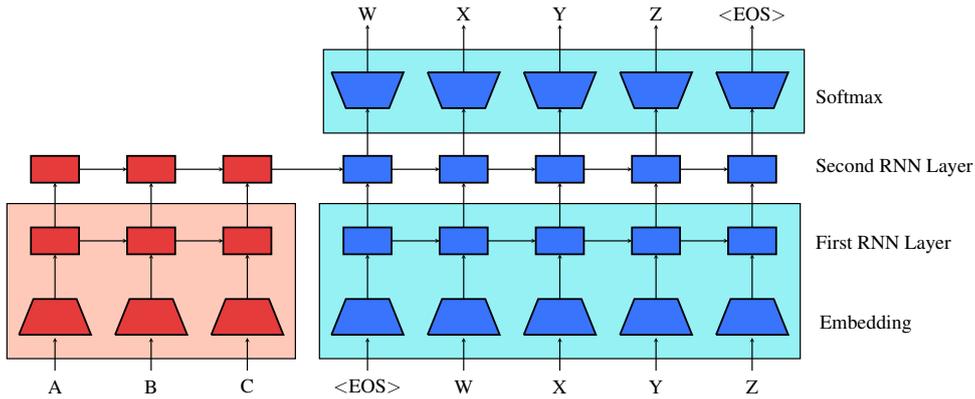}}
\end{center}
\caption{Pretrained sequence to sequence model. The red
  parameters are the encoder and the blue parameters are the
  decoder. All parameters in a shaded box are pretrained, either from
  the source side (light red) or target side (light blue) language model. Otherwise, they are
  randomly initialized.}
\label{figure:architecture}
\end{figure*}

Sequence to sequence (\textit{seq2seq}) models
\cite{SequenceToSequence,cho2014learning,kalchbrenner2013recurrent,Allen1987,Neco1997}
are extremely effective on a variety of tasks that require a mapping
between a variable-length input sequence to a variable-length output
sequence.
The main weakness of sequence to sequence models, and deep networks in
general, lies in the fact that they can easily overfit when the amount
of supervised training data is small.

In this work, we propose a simple and effective technique for using
unsupervised pretraining to improve seq2seq models. Our proposal is to
initialize both encoder and decoder networks with pretrained weights
of two language models. These pretrained weights are then fine-tuned
with the labeled corpus. During the fine-tuning phase, we jointly train the seq2seq objective with the language modeling objectives to prevent overfitting.

We benchmark this method on machine translation for
English$\rightarrow$German and abstractive summarization on CNN and
Daily Mail articles.  Our
main result is that a seq2seq model, with pretraining, exceeds the
strongest possible baseline in both neural machine translation and
phrase-based machine translation. Our model obtains an improvement of
1.3 BLEU from the previous best models on both WMT'14 and WMT'15
English$\rightarrow$German. On human evaluations for abstractive summarization, we find that our model outperforms a purely supervised baseline, both in terms of correctness and in avoiding unwanted repetition.

We also perform ablation studies to understand the behaviors of the
pretraining method. Our study confirms that among many other possible
choices of using a language model in seq2seq with attention, the above
proposal works best.  Our study also shows that, for translation, the
main gains come from the improved generalization due to the pretrained
features. For summarization, pretraining the encoder gives large improvements, suggesting that the gains come from the improved
optimization of the encoder that has been unrolled
for hundreds of timesteps.  On both tasks, our proposed method always
improves generalization on the test sets.

\section{Methods}

In the following section, we will describe our basic unsupervised
pretraining procedure for sequence to sequence learning and how to
modify sequence to sequence learning to effectively make use of the
pretrained weights. We then show several extensions to improve the
basic model.

\subsection{Basic Procedure}

Given an input sequence $x_1, x_2, ..., x_{m}$ and an output sequence
$y_n, y_{n-1}, ..., y_1$, the objective of sequence to sequence learning
is to maximize the likelihood $p(y_n, y_{n-1}, ..., y_1 | x_1, x_2,
..., x_{m})$. Common sequence to sequence learning methods decompose this objective as
$p(y_n, y_{n-1}, ..., y_1 |
x_1, x_2, ..., x_{m}) = \prod_{t=1}^n p(y_t| y_{t-1}, ..., y_1; x_1, x_2, ...,
x_{m})$.

In sequence to sequence learning, an RNN encoder is used to represent
$x_1, ..., x_m$ as a hidden vector, which is given to an RNN decoder
to produce the output sequence. Our method is based on the observation
that without the encoder, the decoder essentially acts like a language
model on $y$'s. Similarly, the encoder with an additional output layer also acts like a language
model. Thus it is natural to use trained languages models to initialize the encoder and decoder.

Therefore, the basic procedure of our approach is to pretrain both the
seq2seq encoder and decoder networks
with language models, which can be trained on large amounts of
unlabeled text data. This can be seen in
Figure~\ref{figure:architecture}, where the parameters in the shaded
boxes are pretrained. In the following we will describe the method in
detail using machine translation as an example application.

First, two monolingual datasets are collected, one for the source side
language, and one for the target side language. A language model (\textit{LM})
is trained on each dataset independently, giving an LM trained on the
source side corpus and an LM trained on the target side corpus.

After two language models are trained, a multi-layer seq2seq model $M$
is constructed. The embedding and first LSTM layers of the encoder and
decoder are initialized with the pretrained weights. To be even more
efficient, the softmax of the decoder is initialized with the
softmax of the pretrained target side LM.

\subsection{Monolingual language modeling losses}

After the seq2seq model $M$ is initialized with the two LMs, it is
fine-tuned with a labeled dataset. However, this procedure may lead to \emph{catastrophic forgetting}, where the model's performance on the language modeling tasks falls dramatically after fine-tuning  \citep{goodfellow2013empirical}. This may hamper the model's ability to generalize, especially when trained on small labeled datasets.

To ensure that the model does not
overfit the labeled data, we regularize the parameters that were pretrained
by continuing to train with the monolingual language modeling losses.
The seq2seq and language modeling losses are weighted equally.

In our ablation study, we find that this technique is complementary to pretraining and is important in achieving high performance.

\subsection{Other improvements to the model}
\begin{figure}[h]
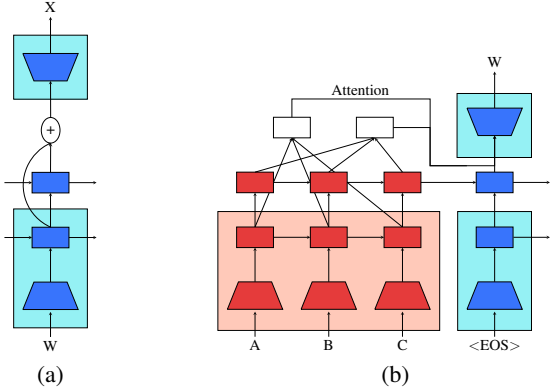

\begin{center}
%\subfloat[]{\scalebox{0.3}{\input{images/lmcost.tex}}}
%\qquad \qquad
\subfloat[]{\scalebox{0.23}{\input{images/residual_connection_2.tex}}}
\qquad \qquad
\subfloat[]{\scalebox{0.23}{\input{images/attention.tex}}}
\end{center}
\caption{Two small improvements to the baseline model: (a) residual connection, and (b) multi-layer attention.}
\label{figure:improvements}
\end{figure}

Pretraining and the monolingual language modeling losses provide the vast majority of improvements to the model. However in early experimentation, we found minor but consistent improvements with two additional techniques: a) residual connections and b) multi-layer attention (see
Figure~\ref{figure:improvements}).

\paragraph{Residual connections:} As described, the input vector to the decoder softmax layer is a
random vector because the high level (non-first) layers of the LSTM
are randomly initialized. This introduces
random gradients to the pretrained parameters. To avoid this, we use a
residual connection from the output of the first LSTM layer directly
to the input of the softmax (see Figure~\ref{figure:improvements}-a).

\paragraph{Multi-layer attention:} In all our models, we use an attention mechanism \citep{Attention}, where the model attends over both top and first layer (see Figure~\ref{figure:improvements}-b).
More concretely, given a query vector $q_t$ from the decoder, encoder
states from the first layer $h_1^1, \ldots, h_T^1$, and encoder states
from the last layer $h_1^L, \ldots, h_T^L$, we compute the attention
context vector $c_t$ as follows:
\begin{align*}
	\alpha_i &= \frac{\exp(q_t \cdot h_i^N)}{\sum_{j=1}^T \exp(q_t \cdot h_j^N)} &
    c_t^1 &= \sum_{i=1}^{T} \alpha_i h_i^1 & \\
	c_t^N &= \sum_{i=1}^{T} \alpha_i h_i^N &
    c_t &= [c_t^1;c_t^N]
\end{align*}

\section{Experiments}
In the following section, we apply our approach to two important tasks in
seq2seq learning: machine translation and abstractive
summarization. On each task, we
compare against the previous best systems. We also perform ablation
experiments to understand the behavior of each component of our
method.

\subsection{Machine Translation}

\begin{table*}[h!]
  \small
  \begin{center}
\begin{tabular}{lc|cc}
& & \multicolumn{2}{c}{\textit{BLEU}} \\
\textit{System} & \textit{ensemble?} & \textit{newstest2014} & \textit{newstest2015} \\ \hline 
Phrase Based MT \citep{William2016} & - & 21.9 & 23.7 \\ \hline
Supervised NMT \citep{Jean} & single & - & 22.4 \\
Edit Distance Transducer NMT \citep{Stahlberg2016} & single & 21.7 & 24.1 \\
Edit Distance Transducer NMT \citep{Stahlberg2016} & ensemble 8 & 22.9 & 25.7 \\\hline
Backtranslation \citep{Backtranslation} & single & 22.7 & 25.7 \\
Backtranslation \citep{Backtranslation} & ensemble 4  & 23.8 & 26.5 \\
Backtranslation \citep{Backtranslation} & ensemble 12 & \textbf{24.7} & 27.6 \\ \hline
No pretraining & single & 21.3 & 24.3 \\
Pretrained seq2seq & single & \textbf{24.0} & \textbf{27.0} \\
Pretrained seq2seq & ensemble 5 & \textbf{24.7} & \textbf{28.1}
\end{tabular}
\end{center}
\caption{English$\rightarrow$German performance on WMT test sets. Our pretrained model outperforms all other models. Note that the model without pretraining uses the LM objective.}
\label{table:mt-sota}
\end{table*}

\begin{figure*}[h!]
\begin{center}
\includegraphics[scale=0.65]{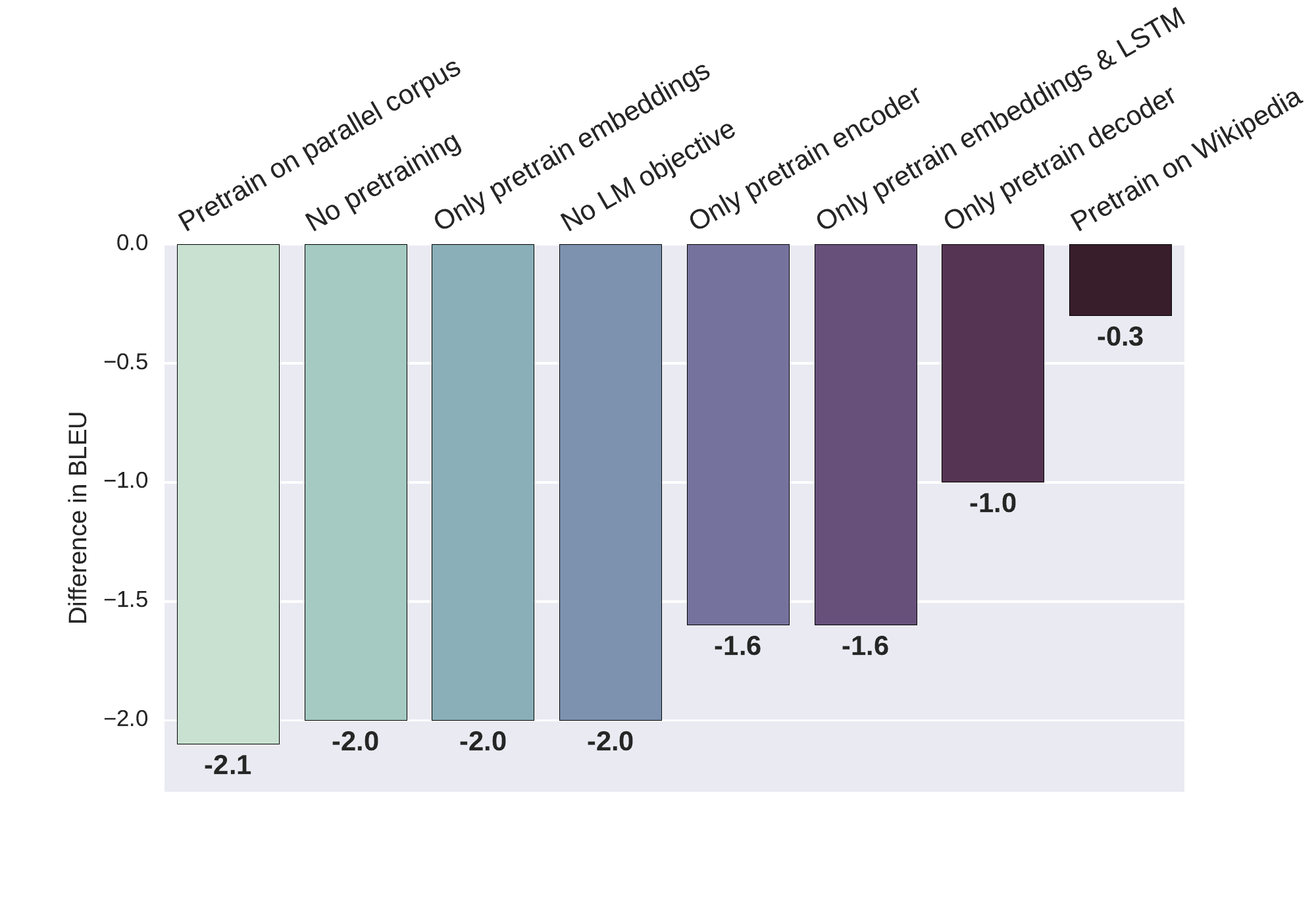}
\caption{English$\rightarrow$German ablation study measuring the difference in validation BLEU between various ablations and the full model. More negative is worse. The full model uses LMs trained with monolingual data to initialize the encoder and decoder, plus the language
modeling objective.}
\label{figure:mt-ablation}
\end{center}
\end{figure*}

\paragraph{Dataset and Evaluation:} For machine translation, we evaluate our method on the WMT
English$\rightarrow$German task \citep{WMT2015}. We used the WMT 14
training dataset, which is slightly smaller than the WMT 15
dataset. Because the dataset has some noisy examples, we used a language detection system to filter the training examples.
Sentences pairs where either the source was not English or the target
was not German were thrown away. This resulted in
around 4 million training examples. Following \citet{Backtranslation}, we use subword units \citep{BPE} with 89500
merge operations, giving a vocabulary size around 90000. The validation set is the concatenated newstest2012 and
newstest2013, and our test sets are newstest2014 and
newstest2015. Evaluation on the validation set was with case-sensitive
BLEU \citep{BLEU} on tokenized text using
\texttt{multi-bleu.perl}. Evaluation on the test sets was with
case-sensitive BLEU on detokenized text using \texttt{mteval-v13a.pl}.
The monolingual training datasets are the News Crawl English and
German corpora, each of which has more than a billion tokens.

\paragraph{Experimental settings:} The language models were trained in the same fashion as
\citep{ExploringTheLimitsOfLanguageModeling} We used a 1 layer 4096 dimensional LSTM with the hidden state projected down to 1024 units \citep{LSTMProjection} and trained for one
week on 32 Tesla K40 GPUs. Our seq2seq model was a 3 layer model, where
the second and third layers each have 1000 hidden units. The monolingual
objectives, residual connection, and the modified attention were all
used. We used the Adam optimizer \citep{Adam} and train with asynchronous
SGD on 16 GPUs for speed. We used a learning rate of 5e-5 which is
multiplied by 0.8 every 50K steps after an initial 400K steps, gradient clipping with norm 5.0 \citep{GradientClipping}, and dropout of 0.2 on non-recurrent connections
\citep{RNNDropout}. We used early stopping on validation set perplexity. A beam size of 10 was used for decoding. Our
ensemble is constructed with the 5 best performing models on the
validation set, which are trained with different hyperparameters.

\paragraph{Results:} Table \ref{table:mt-sota} shows the results of our method in
comparison with other baselines. Our method achieves a new
state-of-the-art for single model performance on both newstest2014 and
newstest2015, significantly outperforming the competitive semi-supervised \textit{backtranslation} technique \citep{Backtranslation}.
Equally impressive is the fact that our best single
model outperforms the previous state of the art ensemble of 4
models. Our ensemble of 5 models matches or exceeds the previous best
ensemble of 12 models.

\paragraph{Ablation study:} In order to better understand the effects of pretraining, we conducted
an ablation study by modifying the pretraining scheme. We were primarily interested in varying the pretraining scheme and the monolingual language modeling objectives because these two techniques produce the largest gains in the model. Figure
\ref{figure:mt-ablation} shows the drop in validation BLEU of various
ablations compared with the full model. The \textit{full model} uses LMs trained with monolingual data to initialize the encoder and decoder, in addition to the language
modeling objective. In the following, we interpret the findings of the study. Note that some findings are specific to the translation task.

Given the results from the ablation study, we can make the following
observations:
\begin{itemize}
\item Only pretraining the decoder is better than only pretraining the encoder: Only pretraining the encoder
leads to a 1.6 BLEU point drop while only pretraining the decoder leads to a 1.0 BLEU point drop.
\item Pretrain as much as possible because the benefits compound:
  given the drops of no pretraining at all ($-2.0$) and only
  pretraining the encoder ($-1.6$), the additive estimate of the drop
  of only pretraining the decoder side is $-2.0 - (-1.6) =
  -0.4$; however the actual drop is $-1.0$ which is a much larger drop
  than the additive estimate.
\item Pretraining the softmax is important: Pretraining only the
  embeddings and first LSTM layer gives a large drop of 1.6 BLEU
  points.
\item The language modeling objective is a strong regularizer: The
  drop in BLEU points of pretraining the entire model and not using
  the LM objective is as bad as using the LM objective without
  pretraining.
\item Pretraining on a lot of unlabeled data is essential for learning
  to extract powerful features: If the model is initialized with LMs
  that are pretrained on the source part and target part of the
  \textit{parallel} corpus, the drop in performance is as large as not
  pretraining at all. However, performance remains strong
  when pretrained on the large, non-news Wikipedia corpus.
\end{itemize}

\begin{figure}[h!]
\begin{center}
\includegraphics[scale=0.5]{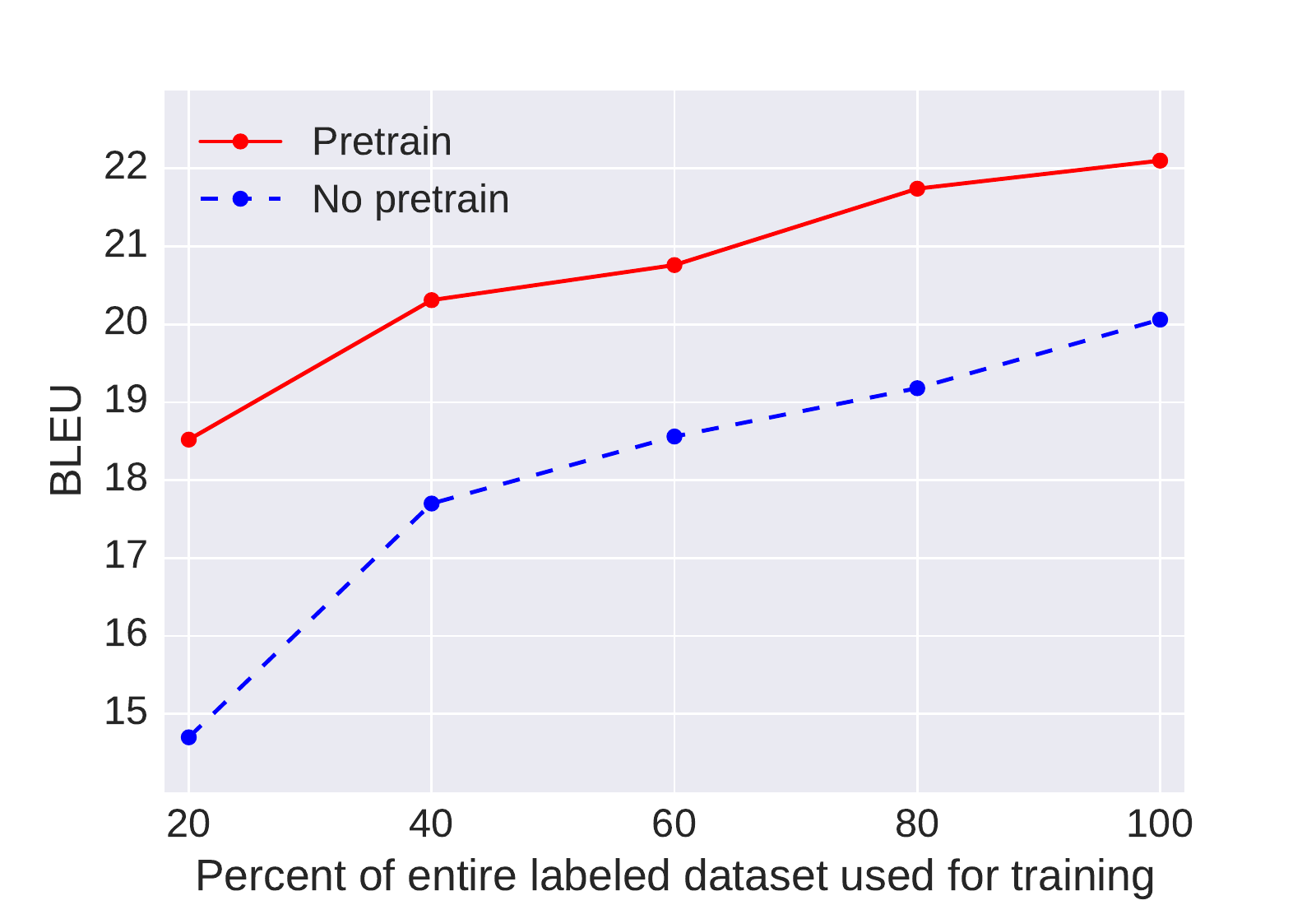}
\caption{Validation performance of pretraining vs. no pretraining when trained on a subset of the entire labeled dataset for English$\rightarrow$German translation.}
\label{figure:percent-of-data}
\end{center}
\end{figure}

To understand the contributions of unsupervised pretraining vs.
supervised training, we track the performance of pretraining as a
function of dataset size. For this, we trained a a model with and
without pretraining on random subsets of the
English$\rightarrow$German corpus. Both models use the additional LM
objective. The results are summarized in Figure
\ref{figure:percent-of-data}. When a 100\% of the labeled data is
used, the gap between the pretrained and no pretrain model is 2.0 BLEU
points. However, that gap grows when less data is available. When
trained on 20\% of the labeled data, the gap becomes 3.8 BLEU
points. This demonstrates that the pretrained models degrade less as
the labeled dataset becomes smaller.

\subsection{Abstractive Summarization}

\begin{table*}[h!]
\begin{center}
\begin{tabular}{l|c|c|c}
\textit{System} & \textit{ROUGE-1} & \textit{ROUGE-2} & \textit{ROUGE-L} \\ \hline
Seq2seq + pretrained embeddings \citep{Ramesh} & 32.49 & 11.84 & 29.47 \\
+ temporal attention \citep{Ramesh} & \textbf{35.46} & \textbf{13.30} & \textbf{32.65} \\ \hline
Pretrained seq2seq & 32.56 & 11.89 & 29.44
\end{tabular}
\caption{Results on the anonymized CNN/Daily Mail dataset.}
\label{table:summ-results}
\end{center}
\end{table*}

\begin{figure*}[h!]
\begin{center}
\includegraphics[scale=0.65]{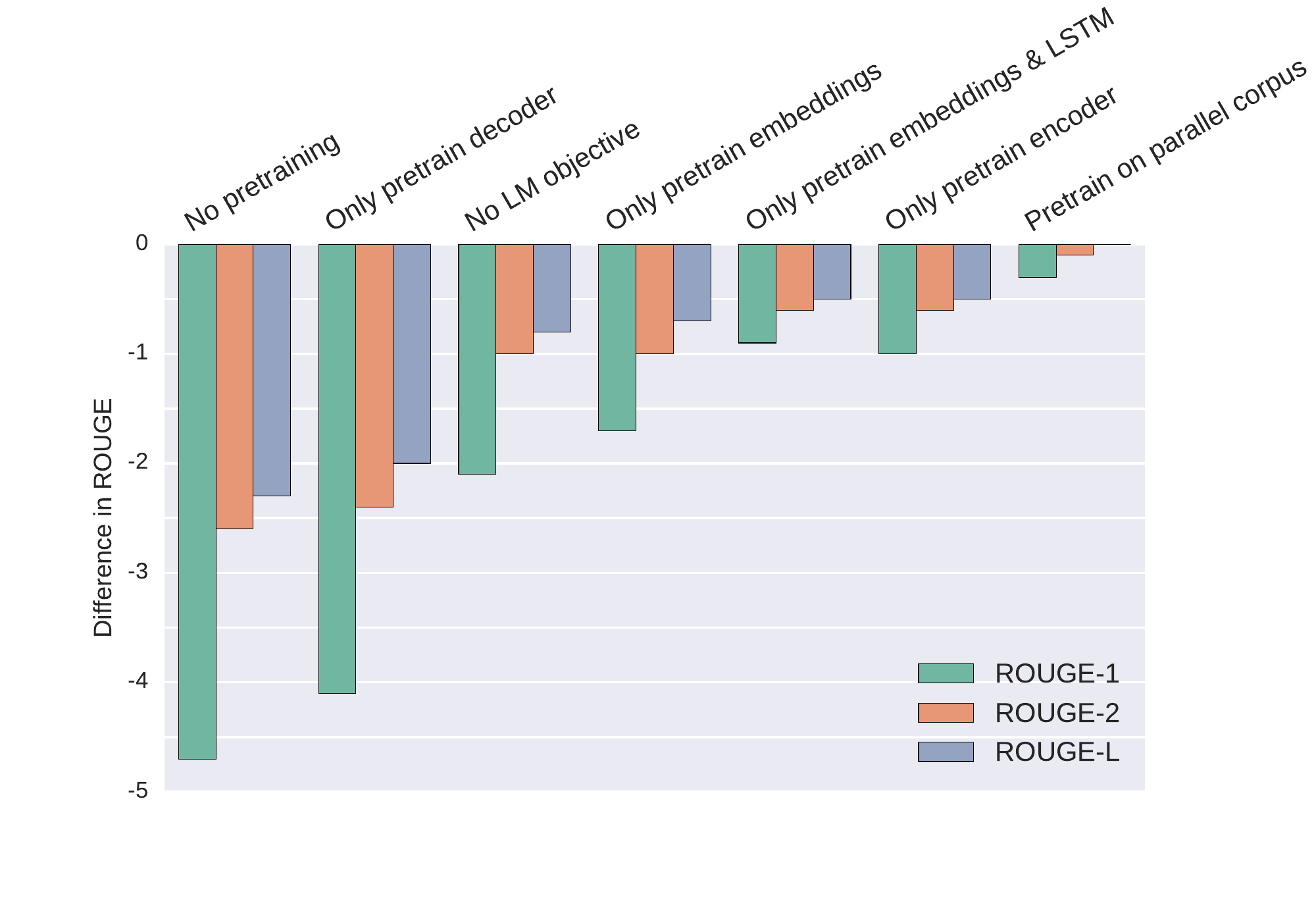}
\caption{Summarization ablation study measuring the difference in validation ROUGE between various ablations and the full model. More negative is worse. The full model uses LMs trained with unlabeled data to initialize the encoder and decoder, plus the language
modeling objective.}
\label{figure:summ-ablation}
\end{center}
\end{figure*}

\paragraph{Dataset and Evaluation:} For a low-resource abstractive summarization task,
we use the CNN/Daily Mail corpus
from \citep{TeachingMachines}. Following \citet{Ramesh}, we modify
the data collection scripts to restore the bullet point summaries. The
task is to predict the bullet point summaries from a news article. The dataset has fewer than 300K document-summary pairs. To
compare against \citet{Ramesh}, we used the anonymized
corpus. However, for our ablation study, we used the non-anonymized
corpus.\footnote{We encourage future researchers to use the non-anonymized
version because it is a more realistic summarization setting with a larger vocabulary. Our numbers on the non-anonymized test set are $35.56$ ROUGE-1, $14.60$ ROUGE-2, and $25.08$ ROUGE-L. We did not consider highlights as separate sentences.}  We evaluate our system using full length ROUGE
\citep{ROUGE}. For the anonymized
corpus in particular, we considered each highlight as a separate
sentence following \citet{Ramesh}. In this setting, we used the English Gigaword corpus \citep{Gigaword}
as our larger, unlabeled ``monolingual'' corpus, although all data used in this task is
in English.

\paragraph{Experimental settings:} We use subword units \citep{BPE} with 31500 merges, resulting in a
vocabulary size of about 32000. We use up to the first 600 tokens of
the document and predict the entire summary.   Only one language model is trained and it is used to
initialize both the encoder and decoder,
since the source and target languages are the same. However, the encoder and
decoder are not tied. The LM is a one-layer LSTM of size 1024 trained in a
similar fashion to \citet{ExploringTheLimitsOfLanguageModeling}. For
the seq2seq model, we use the same settings as the machine translation
experiments. The only differences are that we use a 2 layer model with
the second layer having 1024 hidden units, and that the learning rate
is multiplied by 0.8 every 30K steps after an initial 100K steps.

\paragraph{Results:} Table \ref{table:summ-results} summarizes our results on the
anonymized version of the corpus. Our pretrained model is only able to
match the previous baseline seq2seq of \citet{Ramesh}. Interestingly, they use pretrained word2vec \citep{Word2Vec}
vectors to initialize their word embeddings. As we show in our
ablation study, just pretraining the embeddings itself gives a large improvement. Furthermore, our
model is a unidirectional LSTM while they use a bidirectional
LSTM. They also use a longer context of 800 tokens, whereas we used a context of 600 tokens due to GPU memory issues.

\paragraph{Ablation study:} We
performed an ablation study similar to the one performed on the
machine translation model. The results are reported in Figure \ref{figure:summ-ablation}. Here we
report the drops on ROUGE-1, ROUGE-2, and ROUGE-L on the
non-anonymized validation set.

Given the results from our ablation study, we can make the following
observations:
\begin{itemize}
  \item Pretraining appears to improve optimization: in contrast with the
    machine translation model, it is more beneficial to only pretrain
    the encoder than only the decoder of the summarization model. One
    interpretation is that pretraining enables the gradient to flow
    much further back in time than randomly initialized weights.
    This may also explain why pretraining on the parallel corpus is no worse than
    pretraining on a larger monolingual corpus. 
  \item The language modeling objective is a strong regularizer: A
    model without the LM objective has a significant drop in ROUGE
    scores.
\end{itemize}

\paragraph{Human evaluation:} As ROUGE may not be able to capture the quality of summarization, we
also performed a small qualitative study to understand the human
impression of the summaries produced by different models. We took 200
random documents and compared the performance of a pretrained and
non-pretrained system. The document, gold summary, and the two system
outputs were presented to a human evaluator who was asked to rate
each system output on a scale of 1-5 with 5 being the best score. The
system outputs were presented in random order and the evaluator did
not know the identity of either output. The evaluator noted if
there were repetitive phrases or sentences in either system outputs. Unwanted repetition was also noticed by \citet{Ramesh}.

Table \ref{table:summ-human-score} and \ref{table:summ-human-repeat} show the results of the study. In both cases, the pretrained system outperforms the system without pretraining in a statistically significant manner. The better optimization enabled by pretraining improves the generated summaries and decreases unwanted repetition in the output.

\begin{table}[h!]
\begin{center}
\begin{tabular}{c|c|c}
\textit{NP $>$ P} & \textit{NP = P} & \textit{NP $<$ P} \\ \hline
29 & 88 & 83
\end{tabular}
\caption{The count of how often the no pretrain system (\textit{NP}) achieves a higher, equal, and lower score than the pretrained system (\textit{P}) in the side-by-side study where the human evaluator gave each system a score from 1-5. The sign statistical test gives a p-value of $<0.0001$ for rejecting the null hypothesis that there is no difference in the score obtained by either system.}
\label{table:summ-human-score}
\end{center}
\end{table}

\begin{table}[h!]
\begin{center}
\begin{tabular}{rr|cc}
& & \multicolumn{2}{c}{\textit{No pretrain}} \\
& & No repeats & Repeats \\ \hline
\multirow{2}{*}{\textit{Pretrain}} & No repeats & 67 & 65 \\
						  	 & Repeats & 24 & 44
\end{tabular}
\caption{The count of how often the pretrain and no pretrain systems contain repeated phrases or sentences in their outputs in the side-by-side study. McNemar's test gives a p-value of $< 0.0001$ for rejecting the null hypothesis that the two systems repeat the same proportion of times. The pretrained system clearly repeats less than the system without pretraining.}
\label{table:summ-human-repeat}
\end{center}
\end{table}

\section{Related Work}
Unsupervised pretraining has been intensively studied in the past
years, most notably is the work by \citet{Dahl2012} who found that
pretraining with deep belief networks improved feedforward acoustic
models. More recent acoustic models have found pretraining unnecessary
\citep{Xiong2016,YuZhang2016,chan2015listen}, probably because the
reconstruction objective of deep belief networks is too easy. In
contrast, we find that pretraining language models by next step
prediction significantly improves seq2seq on challenging real world
datasets. 

Despite its appeal, unsupervised learning has not been widely used to
improve supervised training. \citet{DaiSSSL,radford2017} are amongst
the rare studies which showed the benefits of pretraining in a
semi-supervised learning setting. Their methods are similar to ours
except that they did not have a decoder network and thus could not
apply to seq2seq learning. Similarly, \citet{zhangexploiting} found it
useful to add an additional task of sentence reordering of source-side
monolingual data for neural machine translation. Various forms of
transfer or multitask learning with seq2seq framework also have the
flavors of our algorithm \citep{Zoph,MultiTaskSeq2Seq,Firat_ZRNMT}.

Perhaps most closely related to our method is the work by
\citet{DeepFusion}, who combined a language model with an already
trained seq2seq model by fine-tuning additional deep output
layers. Empirically, their method produces small improvements over the
supervised baseline. We suspect that their method
does not produce significant gains because (i) the models are trained
independently of each other and are not fine-tuned (ii) the LM is
combined with the seq2seq model after the last layer, wasting the
benefit of the low level LM features, and (iii) only using the LM on
the decoder side. \citet{venugopalan2016improving} addressed (i) but
still experienced minor improvements. Using pretrained GloVe
embedding vectors \citep{pennington2014glove} had more impact.

Related to our approach in principle is the work by \citet{Chen2016}
who proposed a two-term, theoretically motivated unsupervised
objective for unpaired input-output samples. Though they did not apply
their method to seq2seq learning, their framework can be modified to do so. In that
case, the first term pushes the output to be highly probable under
some scoring model, and the second term ensures that the output
depends on the input. In the seq2seq setting, we interpret the first
term as a pretrained language model scoring the output sequence. In
our work, we fold the pretrained language model into the decoder. We
believe that using the pretrained language model only for scoring is
less efficient that using all the pretrained weights. Our use of
labeled examples satisfies the second term. These connections provide
a theoretical grounding for our work.

In our experiments, we benchmark our method on machine translation,
where other unsupervised methods are shown to give promising
results~\citep{Backtranslation,SemiSupervisedNMT}. In
backtranslation~\citep{Backtranslation}, the trained model is used to
decode unlabeled data to yield extra labeled data. One can argue that
this method may not have a natural analogue to other tasks such as
summarization.  We note that their technique is complementary to ours,
and may lead to additional gains in machine translation. The method of
using autoencoders in \citet{SemiSupervisedNMT} is promising, though it
can be argued that autoencoding is an easy objective and language
modeling may force the unsupervised models to learn better features.

\section{Conclusion}

We presented a novel unsupervised pretraining method to improve
sequence to sequence learning. The method can aid in both
generalization and optimization. Our scheme involves pretraining two
language models in the source and target domain, and initializing the
embeddings, first LSTM layers, and softmax of a sequence to sequence
model with the weights of the language models. Using our method, we
achieved state-of-the-art machine translation results on both WMT'14
and WMT'15 English to German. A key advantage of this technique is that it is flexible and can be
applied to a large variety of tasks.

\bibliography{emnlp2017}
\bibliographystyle{emnlp_natbib}
\appendix
\newpage
\section{Example outputs}
\label{sec:supplemental}

\begin{table*}[h!]
\begin{center}
\begin{tabular}{|L|}
\hline \textbf{Source Document} \\ \hline
( cnn ) like phone booths and typewriters , record stores are a vanishing breed -- another victim of the digital age . camelot music . virgin megastores . wherehouse music . tower records . all of them gone . corporate america has largely abandoned brick - and - mortar music retailing to a scattering of independent stores , many of them in scruffy urban neighborhoods . and that s not necessarily a bad thing . yes , it s harder in the spotify era to find a place to go buy physical music . but many of the remaining record stores are succeeding -- even thriving -- by catering to a passionate core of customers and collectors . on saturday , hundreds of music retailers will hold events to commemorate record store day , an annual celebration of , well , your neighborhood record store . many stores will host live performances , drawings , book signings , special sales of rare or autographed vinyl and other happenings . some will even serve beer . to their diehard customers , these places are more than mere stores : they are cultural institutions that celebrate music history ( the entire duran duran oeuvre , all in one place ! ) , display artifacts ( aretha franklin on vinyl ! ) , and nurture the local music scene ( hey , here s a cd by your brother s metal band ! ) . they also employ knowledgeable clerks who will be happy to debate the relative merits of blood on the tracks and blonde on blonde . or maybe , like jack black in high fidelity , just mock your lousy taste in music . so if you re a music geek , drop by . but you might think twice before asking if they stock i just called to say i love you .
\\ \hline
 \textbf{Ground Truth summary} \\ \hline
saturday is record store day , celebrated at music stores around the world . many stores will host live performances , drawings and special sales of rare vinyl .
\\ \hline
\textbf{No pretrain} \\ \hline
corporate america has largely abandoned brick - brick - mortar music . many of the remaining record stores are succeeding -- even thriving -- by catering to a passionate core of customers .
\\ \hline
\textbf{Pretrained} \\ \hline
hundreds of music retailers will hold events to commemorate record store day . many stores will host live performances , drawings , book signings , special sales of rare or autographed vinyl .
\\ \hline
\end{tabular}
\caption{The pretrained model outputs a highly informative summary, while the no pretrain model outputs irrelevant details.}
\label{table:summ-example1}
\end{center}
\end{table*}

\begin{table*}[h!]
\begin{center}
\begin{tabular}{|L|}
\hline \textbf{Source Document} \\ \hline
 ( cnn ) hey , look what i did . that small boast on social media can trigger a whirlwind that spins into real - life grief , as a texas veterinarian found out after shooting a cat . dr. kristen lindsey allegedly shot an arrow into the back of an orange tabby s head and posted a proud photo this week on facebook of herself smiling , as she dangled its limp body by the arrow s shaft . lindsey added a comment , cnn affiliate kbtx reported . my first bow kill , lol . the only good feral tomcat is one with an arrow through it s head ! vet of the year award ... gladly accepted . callers rang the phones hot at washington county s animal clinic , where lindsey worked , to vent their outrage . web traffic crashed its website . high price of public shaming on the internet then an animal rescuer said that lindsey s prey was probably not a feral cat but the pet of an elderly couple , who called him tiger . he had gone missing on wednesday , the same day that lindsey posted the photo of the slain cat . cnn has not been able to confirm the claim . as the firestorm grew , lindsey wrote in the comments underneath her post : no i did not lose my job . lol . psshh . like someone would get rid of me . i m awesome ! that prediction was wrong . the clinic fired lindsey , covered her name on its marquee with duct tape , and publicly distanced itself from her actions . our goal now is to go on and try to fix our black eye and hope that people are reasonable and understand that those actions do nt anyway portray what we re for here at washington animal clinic , said dr. bruce buenger . we put our heart and soul into this place . the clinic told wbtx that lindsey was not available for comment . cnn is reaching out to her . she removed her controversial post then eventually shut down her facebook page . callers also complained to the brenham police department and washington county animal control , as her facebook post went viral . the sheriff s office in austin county , where the cat was apparently shot , is investigating , and lindsey could face charges . its dispatchers were overloaded with calls , the sheriff posted on facebook . we are asking you to please take it easy on our dispatchers . as soon as the investigation is complete , we will post the relevant information here on this page , the post read . animal rights activists are pushing for charges . animal cruelty must be taken seriously , and the guilty parties should be punished to the fullest extent of the law , said cat advocacy activist becky robinson . her organization , alley cat allies , is offering a \$ 7,500 reward for evidence leading to the arrest and conviction of the person who shot the cat . but others stood up for lindsey . she s amazing . she s caring , said customer shannon stoddard . she s a good vet , so maybe her bad choice of posting something on facebook was not good . but i do nt think she should be judged for it . she dropped off balloons at the animal clinic for lindsey with a thank you note . cnn s jeremy grisham contributed to this report .
 \\ \hline
 \textbf{Ground Truth summary} \\ \hline
dr. kristen lindsey has since removed the post of her holding the dead cat by an arrow . her employer fired her ; the sheriff s office is investigating . activist offers \$ 7,500 reward .
\\ \hline
\textbf{No pretrain} \\ \hline
dr. kristen lindsey allegedly shot an arrow into the back of an orange orange tabby s head . it s the only good good tomcat is one with an arrow through it s head ! vet vet of the year award .
\\ \hline
\textbf{Pretrained} \\ \hline
lindsey lindsey , a texas veterinarian , shot an arrow into the back of an orange tabby s head . she posted a photo of herself smiling , as she dangled its limp body by the arrow s shaft . lindsey could face charges , the sheriff s department says .
\\ \hline
\end{tabular}
\caption{The pretrained model outputs a highly relevant summary but makes a mistake on the feline executioner's name. The no pretrain model degenerates into irrelevant details and repeats itself.}
\label{table:summ-example2}
\end{center}
\end{table*}

\begin{table*}[h!]
\begin{center}
\begin{tabular}{|L|}
\hline \textbf{Source Document} \\ \hline
eugenie bouchard s run of poor form continued as the top seed was beaten 6 - 3 , 6 - 1 by american lauren davis in the second round at the family circle cup in charleston on wednesday . davis , 21 , had lost her only career meeting with bouchard , but was in control this time against the world no 7 . davis won nine of the final 11 games of the match and broke bouchard s serve twice in the final set to pull off the upset . eugenie bouchard fires down a serve during her second - round match at the family circle cup bouchard shows her frustrations during her straight - sets defeat by lauren davis on wednesday i ve never beaten here before , so i came out knowing i had nothing to lose , said davis , ranked 66th in the world . bouchard was a semi-finalist last year but had struggled in the lead - up to the charlston green - clay event , losing three of her last five matches to lower - ranked opponents . davis used precise groundstrokes to keep bouchard on her heels throughout the second set . davis broke bouchard s serve to love to take a 3 - 1 lead as the 21 - year - old canadian had a double fault , then sailed a forehand long to fall behind . two games later , davis hit a backhand to send back bouchard s 102mph serve for a winner down the left sideline and a 5 - 1 lead . davis hits a forehand on her way to an impressive win against the canadian top seed bouchard has struggled recently , this time slumping to her fourth defeat in six matches the match ended on bouchard s forehand into the net , davis waving to the cheering crowd . bouchard said : as soon as it was over , definitely a bit of anger , but also kind of this confusion , slash , like quest to find what s wrong , like this kind of searching feeling that , ok , like i know something s off . i know something s not right . so i want to find it . i was definitely a little bit slow today , overpowered . usually , i m the one dominating . so it was definitely , just , i do nt know , just not good .
 \\ \hline
\textbf{Ground Truth summary} \\ \hline
eugenie bouchard suffered her fourth defeat in six matches . the canadian top seed lost to lauren davis at the family circle cup . world no 66 davis won 6 - 3 , 6 - 1 in the second round in charleston . davis won nine of the final 11 games of the match to seal victory . click here for all the latest news from charleston .
\\ \hline
\textbf{No pretrain} \\ \hline
bouchard beat american lauren davis 6 - 3 , 6 - 1 in the second round of the family circle cup in charleston on wednesday . bouchard had lost her only career meeting with bouchard but was in control this time against the world no 7 . bouchard won nine of the final 11 games of the match and broke bouchard s serve twice in the final set to pull off the upset .
\\ \hline
\textbf{Pretrained} \\ \hline
eugenie bouchard was beaten 6 - 3 , 6 - 1 by american lauren davis in the second round . davis had lost her only career meeting with bouchard , but was in control this time against the world no 7 . davis hit a backhand to send back bouchard s 102mph serve for a winner down the left sideline .
\\ \hline
\end{tabular}
\caption{Both models output a relevant summary, but the no pretrain model uses the same name to refer to both players.}
\label{table:summ-example3}
\end{center}
\end{table*}

\begin{table*}[h!]
\begin{center}
\begin{tabular}{|L|}
\hline \textbf{Source Document} \\ \hline
( cnn ) mike rowe is coming to a river near you . sometimes , you hear about a person who makes you feel good about humanity , but bad about yourself , rowe says . on thursday s episode of somebody s got ta do it , rowe meets up with chad pregracke , the founder of living lands \& waters , who does just that . pregracke wants to clean up the nation s rivers one piece of detritus at a time . his quota ? always more . read mike rowe s facebook post on how to break our litter habit . since he founded the nonprofit in 1998 at the ripe age of 23 , pregracke and more than 87,000 volunteers have collected 8.4 million pounds of trash from u.s. waterways . those efforts helped him earn the 2013 cnn hero of the year award , along with numerous other honors . wherever you are , no matter if there s a stream , a creek , a lake , whatever , that needs to be cleaned up , you can do it . just organize it and do it , he told cnn s anderson cooper after his win . pregracke also gives rowe a tour of the 150 - foot , solar - powered barge that the living lands \& waters staff calls home during lengthy cleanups . the part - home , part - office , part - dumpster has seven bedrooms , two bathrooms , a classroom and a kitchen -- and just happens to be made from a recycled strip club . according to the organization s latest annual report , pregracke has made it his mission in 2015 to remove 500,000 more pounds of trash . if you d like to help achieve this goal , visit his website to learn how to help : livinglandsandwaters.org / get - involved / .
 \\ \hline
\textbf{Ground Truth summary} \\ \hline
chad pregracke was the 2013 cnn hero of the year . mike rowe visited pregracke for an episode of somebody s got ta do it .
\\ \hline
\textbf{No pretrain} \\ \hline
rowe meets up with chad pregracke , founder of living lands \& waters . pregracke and more than 87,000 volunteers collected 8.4 million pounds of trash from u.s. waterways .
\\ \hline
\textbf{Pretrained} \\ \hline
rowe is the founder of living lands \& waters , who does just that . pregracke also gives rowe a tour of the 150 - foot barge that the living lands \& waters gets .
\\ \hline
\end{tabular}
\caption{A failure case. The pretrained model outputs irrelevant details while the no pretrain model successfully summarizes the document.}
\label{table:summ-example4}
\end{center}
\end{table*}

\begin{table*}[h!]
\begin{center}
\begin{tabular}{|L|}
\hline \textbf{Source} \\ \hline
Mayor Bloomberg told reporters that, because of that court order, the city had suspended the reopening of the public space and protesters were informed, however, that local laws do not allow them to re-install with camping shops and sleeping bags.
 \\ \hline
\textbf{Ground Truth} \\ \hline
Bürgermeister Bloomberg stellt vor der Presse klar , das aufgrund dieser richterlichen Anordnung die erneute Öffnung des Platzes für den Publikumsverkehr und die Demonstranten aufgehoben worden sei . Die Demonstranten wies er darauf hin , dass die Stadtgesetze ihnen nicht erlaubten , sich erneut mit Zelten und Schlafsäcken an diesem Ort einzurichten . 
\\ \hline
\textbf{No pretrain} \\ \hline
Der Bürgermeister Bloomberg sagte den Reportern , dass die Stadt aufgrund dieser Gerichtsentscheidung die Wiedereröffnung des öffentlichen Raumes und die Information der Demonstranten ausgesetzt habe , dass die lokalen Gesetze ihnen nicht erlauben , mit den Campingplätzen und Schlafsäcken neu zu installieren . 
\\ \hline
\textbf{Pretrained} \\ \hline
Bürgermeister Bloomberg erklärte gegenüber Journalisten , dass die Stadt aufgrund dieser Gerichtsentscheidung die Wiedereröffnung des öffentlichen Raums ausgesetzt habe und dass die Demonstranten darüber informiert wurden , dass die örtlichen Gesetze es ihnen nicht erlauben würden , sich mit Campingplätzen und Schlafsälen neu zu installieren . 
\\ \hline
\end{tabular}
\caption{The no pretrain model makes a complete mistranslation when outputting "und die Information der Demonstranten ausgesetzt habe". That translates to "the reopening of the public space and the information [noun] of the protesters were suspended", instead of informing the protesters. Furthermore, it wrongly separated the two sentences, so the first sentence has extra words and the second sentence is left without a subject. The pretrained model does not make any of these mistakes. However, both models make a vocabulary mistake of "zu installieren", which is typically only used to refer to installing software. A human evaluator fluent in both German and English said that the pretrained version was better.}
\label{table:mt-example1}
\end{center}
\end{table*}

\begin{table*}[h!]
\begin{center}
\begin{tabular}{|L|}
\hline \textbf{Source} \\ \hline
The low February temperatures, not only did they cause losses of millions for the agricultural sector, but they limited the possibilities of the state economy to grow, causing a contraction of the economic activity in general of 3.6 percent in the first half of the year, mainly supported by the historic fall of 31.16 per cent in agriculture, which affected the dynamics of other economic sectors.
\\ \hline
\textbf{Ground Truth} \\ \hline
Die niedrigen Temperaturen im Februar verursachten nicht nur Verluste in Millionenhöhe in der Landwirtschaft , sondern steckten darüber hinaus dem Wachstum der Staatswirtschaft enge Grenzen und verursachten im ersten Vierteljahr einen allgemeinen Rückgang der Wirtschaftstätigkeit um 3,6 Prozent Dieser geht hauptsächlich auf den historischen Abbau der landwirtschaftlichen Entwicklung um 31,16 Prozent zurück , der sich bremsend auf weitere Wirtschaftssektoren auswirkte . 
\\ \hline
\textbf{No pretrain} \\ \hline
Die niedrigen Temperaturen im Februar führten nicht nur zu Verlusten für die Landwirtschaft , sondern sie beschränkten die Möglichkeiten der staatlichen Wirtschaft , wachsen zu wachsen , wodurch die Wirtschaftstätigkeit insgesamt von 3,6 Prozent in der ersten Hälfte des Jahres , hauptsächlich durch den historischen Rückgang von 31.16 \% in der Landwirtschaft , beeinflusst wurde , was die Dynamik anderer Wirtschaftssektoren betraf . 
 \\ \hline
\textbf{Pretrained} \\ \hline
Die niedrigen Temperaturen im Februar führten nicht nur zu Verlusten von Millionen für den Agrarsektor , sondern beschränkten die Möglichkeiten der Staatswirtschaft , zu wachsen , was zu einer Schrumpfung der Wirtschaftstätigkeit im Allgemeinen von 3,6 Prozent in der ersten Hälfte des Jahres führte , hauptsächlich durch den historischen Einbruch von 316 Prozent in der Landwirtschaft , der die Dynamik anderer Wirtschaftsbereiche beeinflusst hatte . 
\\ \hline
\end{tabular}
\caption{The human evaluator noted that the pretrained version is better, as it correctly captures the meaning and sentence structure of the middle. The no pretrain model does not misses translating the word "million", repeats itself in "wachsen zu wachsen", and puts the verb "beeinflusst wurde" is an unnatural position. However, the pretrained model makes a mistake in the percentage (316\% instead of 31.16\%).}
\label{table:mt-example4}
\end{center}
\end{table*}

\begin{table*}[h!]
\begin{center}
\begin{tabular}{|L|}
\hline \textbf{Source} \\ \hline
To facilitate the inception of the Second World War, they allowed bankers and politicians to create a latent conflict situation by saddling Germany with huge war reparations, thereby making a radicalist example of the impoverished masses, it remained only to introduce a sufficiently convincing culprit and a leader with a simple solution, while also creating a multi-racial Czechoslovakia with a strong German minority to play, and indeed did, the role of a fifth colony, once the war had been ignited.
\\ \hline
\textbf{Ground Truth} \\ \hline
Um den Zweiten Weltkrieg einfacher entfachen zu können , ließen die Banker durch die Politik eine latente Konfliktsituation schaffen , indem sie Deutschland mit gigantischen Kriegsreparationen belegten ; dadurch schufen sie die Voraussetzung verarmter Massen , so dass sie den Deutschen nur noch einen ausreichend starken Führer unterjubeln mussten , der die Schuldigen benannte und einfache Lösungen anbot ; ein weiterer Faktor war die Schaffung des Vielvölkerstaates Tschechoslowakei mit einer starken deutschen Minderheit , die die Funktion einer fünften Kolonne einnehmen sollte und auch einnahm , um den Kriegsbrand zu entfachen . 
\\ \hline
\textbf{No pretrain} \\ \hline
Um die Gründung des Zweiten Weltkriegs zu erleichtern , ermöglichte es den Bankern und Politikern , eine latente Konfliktlage zu schaffen , indem sie Deutschland mit enormen Reparationsforderungen konfrontierte , wodurch ein radikalislamistisches Beispiel der verarmten Massen entstand , es blieb nur , einen ausreichend aussagekräftigen Schuldigen und einen Führer mit einer einfachen Lösung zu etablieren , während gleichzeitig eine multi-ethnische Tschechoslowakei mit einer starken deutschen Minderheit zu spielen war und tatsächlich die Rolle einer fünften Kolonie war . 
 \\ \hline
\textbf{Pretrained} \\ \hline
Um die Einführung des Zweiten Weltkrieges zu erleichtern , ließen sie Banker und Politiker eine latente Konfliktlage schaffen , indem sie Deutschland mit riesigen Reparationszahlungen belieferten , wodurch ein radikalislamistisches Beispiel der verarmten Massen entstand , es blieb nur , einen ausreichend überzeugenden Schuldigen und einen Führer mit einer einfachen Lösung zu präsentieren , während gleichzeitig eine multiethnische Tschechoslowakei mit einer starken deutschen Minderheit geschaffen wurde , um zu spielen , und tatsächlich , die Rolle einer fünften Kolonie , sobald der Krieg entfacht worden war . 
\\ \hline
\end{tabular}
\caption{An example where the English source is poorly worded. Both models output poor translations, but the evaluator noted that the pretrained version is still better than the no pretrain version. Interestingly, both models mistranslate "radical" as "radikalislamistisches", which means "radical Islam", which is probably a bias in the training data.}
\label{table:mt-example3}
\end{center}
\end{table*}

\begin{table*}[h!]
\begin{center}
\begin{tabular}{|L|}
\hline \textbf{Source} \\ \hline
The total vote count will also be done if at the end of the ordinary calculation is established that the difference between the winner and the candidate placed on second position is equal to or less than one percentage point, as long as there is a request of the representative of the political party whose candidate came on the second position, case in which there will be excluded the electoral boxes that have been considered during the partial recount.
\\ \hline
\textbf{Ground Truth} \\ \hline
Die Stimmenauszählung kann auch in ihrer Gesamtheit erfolgen , wenn nach Abschluss der ordentlichen Berechnung festgestellt wird , dass der Unterschied zwischen dem mutmaßlichen Gewinner und dem Kandidaten auf dem zweiten Platz gleich oder geringer als ein Prozent ist , vorausgesetzt es liegt ein ausdrücklicher Antrag von einem Vertreter der Partei , deren Kandidat Zweiter geworden ist , vor . In diesem Fall würden die Wahlpakete , die einer teilweisen Auszählung ausgesetzt wurden , ausgeschlossen .
\\ \hline
\textbf{No pretrain} \\ \hline
Die gesamte Stimmenanzahl wird auch dann erreicht , wenn am Ende der ordentlichen Berechnung festgestellt wird , dass der Unterschied zwischen dem Sieger und dem Kandidaten , der auf der zweiten Position liegt , gleich oder weniger als einen Prozentpunkt beträgt , vorausgesetzt , dass der Vertreter der Partei , deren Kandidat auf der zweiten Position ist , der Fall ist , in dem die Wahlunterlagen , die während der teilweisen Rückzählung berücksichtigt wurden , ausgeschlossen werden . 
 \\ \hline
\textbf{Pretrained} \\ \hline
Die Gesamtzahl der Stimmzettel wird auch dann durchgeführt , wenn am Ende der ordentlichen Berechnung festgestellt wird , dass der Unterschied zwischen dem Gewinner und dem auf den zweiten Platz platzierten Kandidaten gleich oder weniger als einen Prozentpunkt beträgt , solange es einen Antrag des Vertreters der politischen Partei gibt , dessen Kandidat auf die zweite Position kam , in dem es die Wahlzettel ausklammert , die während der Teilzählung berücksichtigt wurden . 
\\ \hline
\end{tabular}
\caption{Another example where the English source is poorly worded. Both models get the structure right, but have a variety of problematic translations. Both models miss the meaning of "total vote count". They both also translate "electoral boxes" poorly - the no pretrain model calls it "electoral paperwork" while the pretrained model calls it "ballots". These failures may be because of the poorly worded English source. The human evaluator found them both equally poor.}
\label{table:mt-example2}
\end{center}
\end{table*}

\end{document}